\documentclass[letterpaper, 10pt, conference]{ieeeconf}      %

\IEEEoverridecommandlockouts                          
\usepackage{times}
\usepackage[utf8]{inputenc} %
\usepackage[T1]{fontenc}    %
\usepackage{url}            %
\usepackage{booktabs}       %
\usepackage{nicefrac}       %
\usepackage{microtype}      %
\usepackage{graphics,color}
\usepackage{arydshln}
\usepackage{algorithm}
\usepackage{mathtools}
\let\labelindent\relax
\usepackage{enumitem}
\usepackage{multirow}
\usepackage{placeins}

\usepackage{amsfonts}       %
\usepackage{amsmath}       %
\usepackage{amssymb}

\newcommand{\Figref}[1]{Figure~\ref{#1}}  %
\newcommand{\figref}[1]{Fig.~\ref{#1}}    %

\newcommand{\tabref}[1]{Tab.~\ref{#1}}
\newcommand{\Eqnref}[1]{Equation~\ref{#1}}
\newcommand{\eqnref}[1]{Eq.~\ref{#1}} %
\newcommand{\secref}[1]{Sec.~\ref{#1}} %

\usepackage{xspace}
\makeatletter
\DeclareRobustCommand\onedot{\futurelet\@let@token\@onedot}
\def\@onedot{\ifx\@let@token.\else.\null\fi\xspace}
\makeatother

\usepackage{xr-hyper}
\makeatletter
\newcommand*{\addFileDependency}[1]{%
  \typeout{(#1)}
  \@addtofilelist{#1}
  \IfFileExists{#1}{}{\typeout{No file #1.}}
}
\makeatother

\usepackage{xcolor}
\definecolor{ourblue}{rgb}{0.368,0.507,0.71}
\definecolor{ourorange}{rgb}{0.881,0.611,0.142}
\definecolor{ourgreen}{rgb}{0.56,0.692,0.195}
\definecolor{ourred}{rgb}{0.923,0.386,0.209}
\definecolor{ourviolet}{rgb}{0.528,0.471,0.701}
\definecolor{ourbrown}{rgb}{0.772,0.432,0.102}
\definecolor{ourlightblue}{rgb}{0.364,0.619,0.782}
\definecolor{ourdarkgreen}{rgb}{0.572,0.586,0.}

\definecolor{ourred2}{rgb}{0.84,0.15,0.16}
\definecolor{ourorange2}{rgb}{1,0.5,0.05}
\definecolor{ourblue2}{rgb}{0.12,0.47,0.71}
\definecolor{ourgreen2}{rgb}{0.17,0.63,0.17}
\definecolor{ourpurple3}{rgb}{0.808,0.76,0.88}
\definecolor{ourgreen2}{rgb}{0.29,0.62,0.224}

\usepackage{subcaption}
\graphicspath{../figs/}

\newcommand{\damping}{k_{\mathrm{damp}}\xspace}
\newcommand{\stiffness}{k_{\mathrm{stiff}}\xspace}

\usepackage{pifont}%
\usepackage[switch]{lineno}
\linenumbers

\usepackage{cite}
\usepackage{amsmath,amssymb,amsfonts}
\usepackage{algorithmic}
\usepackage{graphicx}
\usepackage{textcomp}
\usepackage{xcolor}
\usepackage[hidelinks]{hyperref}
\nolinenumbers
\def\BibTeX{{\rm B\kern-.05em{\sc i\kern-.025em b}\kern-.08em
    T\kern-.1667em\lower.7ex\hbox{E}\kern-.125emX}}

\title{\LARGE \bf Learning to Control Emulated Muscles in Real Robots:\\ Towards Exploiting Bio-Inspired Actuator Morphology%

\author{Pierre Schumacher$^{1,2}$, Lorenz Krause$^{1,2}$, Jan Schneider$^{1}$, Dieter Büchler$^{1}$, Georg Martius$^{*1,3}$, Daniel Haeufle$^{*2,4}$}
\thanks{$^{1}$Max Planck Institute for Intelligent Systems, Tübingen, Germany.}
\thanks{$^{2}$Hertie Institute for Clinical Brain Research, and Centre for Integrative Neuroscience, University of Tübingen, Germany.}
\thanks{$^{3}$Distributed Intelligence, University of Tübingen, Germany}
\thanks{$^{4}$Institute of Computer Engineering (ZITI), Heidelberg University, Germany}
\thanks{$^*$GM and DH contributed equally to this publication}
\thanks{The authors thank the International Max Planck Research School for Intelligent Systems (IMPRS-IS) for supporting Pierre Schumacher. This work was supported by the Cyber Valley Research Fund (CyVy-RF-2020-11 to DH and GM)}
}
\begin{document}
\maketitle
\thispagestyle{empty}
\pagestyle{empty}

\begin{abstract}

Recent studies have demonstrated the immense potential of exploiting muscle actuator morphology for natural and robust movement -- in simulation. A validation on real robotic hardware is yet missing. In this study, we emulate muscle actuator properties on hardware in real-time, taking advantage of modern and affordable electric motors. We demonstrate that our setup can emulate a simplified muscle model on a real robot while being controlled by a learned policy. We improve upon an existing muscle model by deriving a damping rule that ensures that the model is not only performant and stable but also tuneable for the real hardware. Our policies are trained by reinforcement learning entirely in simulation, where we show that previously reported benefits of muscles extend to the case of quadruped locomotion and hopping: the learned policies are more robust and exhibit more regular gaits. Finally, we confirm that the learned policies can be executed on real hardware and show that sim-to-real transfer with real-time emulated muscles on a quadruped robot is possible. These results show that artificial muscles can be highly beneficial actuators for future generations of robust legged robots. {Videos: \textcolor{blue}{\url{https://sites.google.com/view/emulatedmuscles}}}

\end{abstract}

\section{Introduction}
\begin{figure*}[htbp]
\centering
\includegraphics[width=0.8\textwidth]{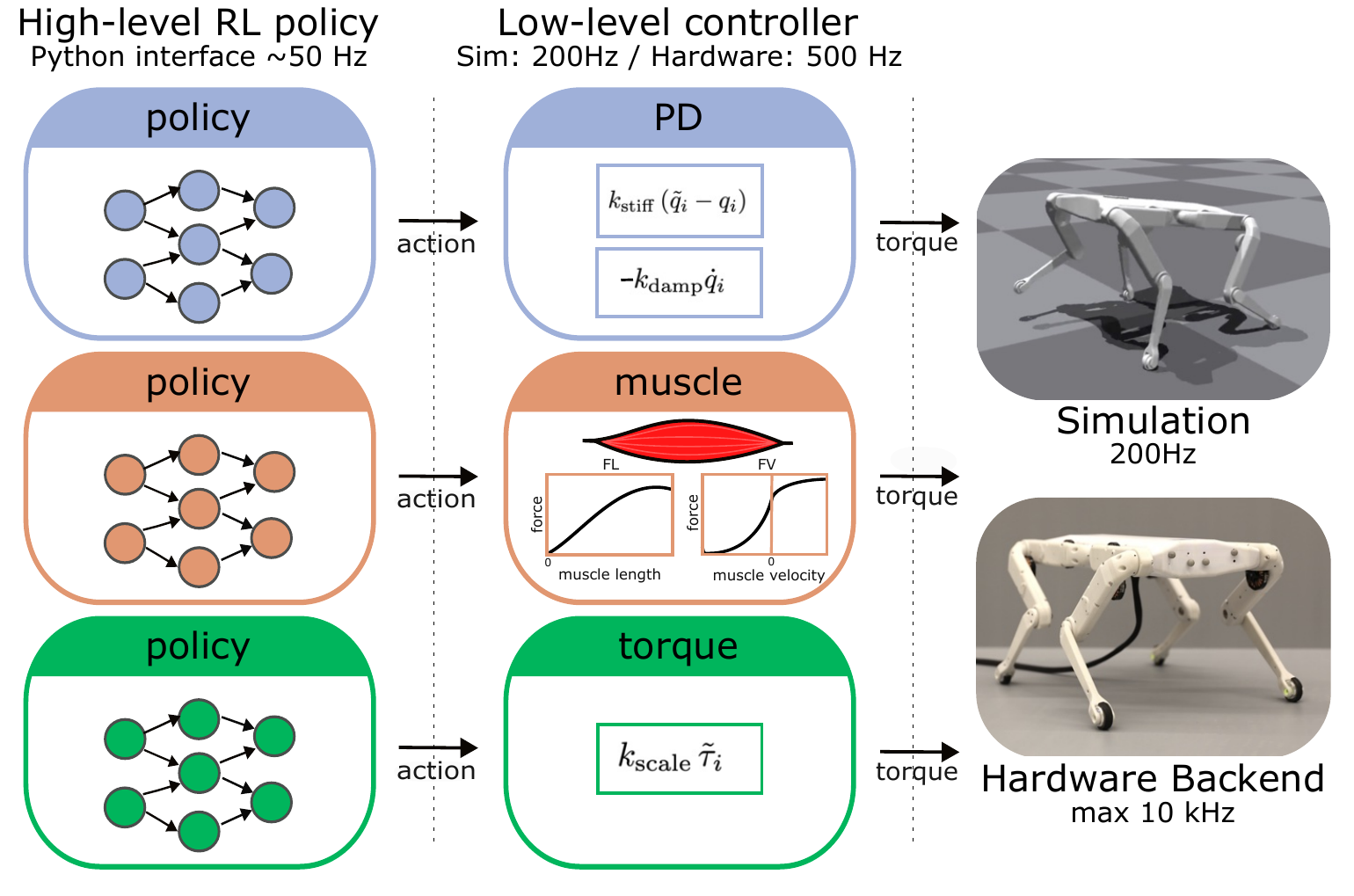}
\caption{\textbf{Emulated muscles on real robots.} We compare three different low-level actuation controllers which are used by three policies trained in simulation: PD, muscle and torque control. The resulting policies can then be tested on a real robot, by a real-time control module running at $\sim 500$ Hz, which communicates with the slowly executed RL policy executed with $\sim 50$ Hz. The robot backend performs torque commands with a maximum of 10 kHz and repeats previous torques between command updates. Importantly, the policy action either represents desired positions, muscle excitations or torques. The simulation has a physics timestep of 5 ms, with 4 physics updates for each controller step.}
\label{fig:overview}
\end{figure*}

The development of learning algorithms for robotic locomotion predominantly relies on position control with electric motors. The ideal actuator is considered to impose as few constraints as possible on the system, while additional properties such as regular foot patterns, minimal torque utilization, and robust policies arise due to the inclusion of a series of cost terms and certain training curricula. Nevertheless, there is rising interest in alternative actuation models, promising advantageous properties amenable to stable control and generalizing controllers without the need for tedious cost term tuning. Some of these studies investigate the effect of action spaces on learned controllers in general~\cite{2017-SCA-action,bohg2019,aljalbout2023role}, while others demonstrate the benefits of a particular actuator class, which is well-known but under-utilized: muscles~\cite{Wochneretal22,vnmc2015,atras2018}. 

Muscles support the nervous system of animals and humans in robust control as they generate zero-time delay reactions (termed \textit{pre}flexes) to environmental influences~\cite{vansoest, Izzi2023, Araz2023}. While some of these advantages have been investigated in simplistic simulated~\cite{Haeufle_2010, 2017-SCA-action, Wochneretal22} and hardware demonstrators~\cite{Seyfarth2007,Mo_2023}, actual real-world experiments remain indispensable to validate the relevance to robotic hardware.

Different artificial muscle architectures are being developed to test the potential of the reported simulation findings, two notable examples being McKibben~\cite{mckibben2006, Ma2023} and HASEL muscles~\cite{hasel2021}. 
Although very promising, many years of research are needed to make these systems reliable and to allow reproducible experimentation in robotics. 
Some approaches aim to speed up the development of control algorithms for artificial muscles by using hybrid sim-hardware setups~\cite{buchler2022learning,Guistetal23} or model-based algorithms~\cite{Ma2023}, which is not feasible for every task.

With the advent of powerful and affordable direct-drive electric motors it is possible to emulate particular actuators in real time. The concept of emulating muscles on motors has been tested in simulation, even for a bipedal robot \cite{vnmc2015, atras2018}, but in hardware only with a singular motor on a lab bench~\cite{dcmuscle2006, Seyfarth2007, dcmuscle_knee2016}. This approach was limited so far, as a PD-controller was required to match the torques predicted by the muscle emulator due to time delays induced by the emulation~\cite{dcmuscle2006, dcmuscle_knee2016}. Benefits of emulated muscles have not been demonstrated in complex and realistic robotic settings. 

Even in those instances where emulated muscles have been simulated, the control algorithms were limited to state machines or reflex-based controllers~\cite{vnmc2015, atras2018}. 
The combination with more powerful frameworks such as reinforcement learning~(RL), is doubly interesting, as it may allow the generation of diverse movements with emulated muscles, while the importance of the actuation model for RL performance is also well documented \cite{2017-SCA-action,aljalbout2023role, schneider2023investigating}, including evidence that muscles can benefit learning~\cite{Wochneretal22}.

The purpose of this study is to demonstrate that emulated muscle properties are beneficial for learning robust walking without the need for excessive reward engineering and can be exploited on real hardware. To achieve this, we first extend simulation results of a previously proposed muscle model~\cite{Wochneretal22}. We implement this model in the GPU-based simulator Isaac Gym~\cite{makoviychuk2021isaac} and show several benefits in quadruped locomotion and hopping tasks --- when compared to commonly used actuators. After observing increased robustness and more regular gaits in the learned policies in simulation, we emulate the muscle-actuator in hardware with a high frequency real-time control cycle without an intermediate PD-controller. Our policies are learned with RL entirely in simulation without the need for numerous fine-tuned cost terms and are shown to transfer to the real system. This finding showcases the effectiveness of muscle actuators and hints at the potential for physical robots, either through emulation or novel hardware actuators.

Contributions: (\textbf{1}) We investigate different actuator models with RL on a simulated robotic quadruped.
(\textbf{2}) We extend previous results on muscle actuation with RL from Wochner et al.~\cite{Wochneretal22} to a realistic robotic locomotion task. (\textbf{3}) We identify a critical parameter of the previously proposed muscle model and derive a theoretically motivated value to facilitate tuning the model for real systems. (\textbf{4}) We achieve sim-to-real transfer with a walking policy to a real-time emulated muscle-controller on robotic hardware.

\section{Methods}

\subsection{Low-level actuation controllers}
We compare the performance of RL policies trained with three different low-level controllers: PD-controller, muscle-controller, and direct torque-controller (Fig.~\ref{fig:overview}).

\textbf{PD-controllers} are commonly used in quadruped locomotion studies~\cite{rudin2021learning, li2022wasabi}; our version receives a desired position computed by the policy, while the desired velocity is set to zero, similar to~\cite{2017-SCA-action}:
\begin{equation}
    \tau_{i} = \stiffness\,(\tilde{q}_{i}-{q}_{i}) - \damping\dot{q}_{i},
\end{equation}
where $\tau_{i}$ is the computed torque, $\tilde{q}_{i}$ the desired position, $q_{i}$ current position, and $\dot{q}_{i}$ the velocity for joint $i$.

The \textbf{torque-controller} is considered to be ideal in the sense that the torque computed by the policy is directly applied to the joint:
\begin{equation}
    \tau_{i} = k_{\mathrm{scale}}\, \tilde{\tau}_{i},
\end{equation}
where $\tilde{\tau}_{i}\in\left[-1, 1\right]$ is computed by the policy and $k_{\mathrm{scale}}$ scales the policy output to the robot's torque range. While it is possible to use torque actuation on quadrupeds with learned policies, it imposes strong control frequency requirements on the implementation~\cite{torque_quadruped}.

The \textbf{muscle-controller} is similar to~\cite{Wochneretal22} and emulates length and velocity dependent muscle force characteristics, as well as activation dynamics:
\begin{equation}
    \tau_{i} = f_{\mathrm{max}}\,\left[\sum_{k=1}^{2} (-1)^{k+1}\mathrm{FL}(l_{k})\,\mathrm{FV}(\bar{\dot{l}}_{k}) \, m_{\mathrm{act},k} + \mathrm{FP}(l_{k})\right],
    \label{eq:muscle}
\end{equation}
where $\mathrm{FL}(\cdot)$ is the force-length and $\mathrm{FV}(\cdot)$ the force-velocity relationship (see ``muscle'' in \figref{fig:overview}), $\mathrm{FP}(\cdot)$ the passive force and $m_{\mathrm{act}}$ the muscle activity. The sum is taken over two muscles for each joint, each pulling in opposing directions. The activity $m_\mathrm{act}$ approaches the policy action $\tilde{a}$ with a low-pass filter:
\begin{equation}
    \dot{m}_\mathrm{act}(t) = \frac{1}{\Delta t_{a}} (\tilde{a}(t) - {m}_\mathrm{act}(t)), 
    \label{eq:act_dyn_mujoco}
\end{equation}
with the time constant $\Delta t_{a}=0.01$\,s.
The scalar $f_{\mathrm{max}}$ was introduced to easily tune the maximum force output. The muscle length is given by:
\begin{equation}
    l_{k} = a_{k}\,q_{i} + b_{k},\label{eq:muscle_len}
\end{equation}
with $a_{k}$ and $b_{k}$ being part of the parametrization. The muscle velocity is given by the derivative of Eq.~\ref{eq:muscle_len}
\begin{equation}
    \bar{\dot{l}}_{k} = \beta\,\dot{l_{k}} = \beta\,(a_{k}\,\dot{q}_{i}).
    \label{eq:muscle_vel}
\end{equation}
Importantly, we multiply the muscle velocity by a scaling parameter $\beta$ in the $\mathrm{FV}$-function (Eq. \ref{eq:muscle_vel}), corresponding to $1/v_\mathrm{max}$ in~\cite{Wochneretal22}, allowing us to tune the maximum damping strength of the muscle. The parameters $a$ and $b$ correspond to $m$ and $l_{\mathrm{ref}}$ in \cite{Wochneretal22}, where the exact formulation is detailed.

\subsection{Muscle model implementation}
For the simulation experiments, we re-implemented the MuJoCo-based muscle model from~\cite{Wochneretal22}. In order to preserve Isaac Gym's computational speed~\cite{rudin2021learning}, we translated the original Cython implementation of the muscle model to a vectorized PyTorch version. Overall, the vanilla PD-controller is only 30\% faster than the muscle implementation, even though we simulate 16 muscles for the robot---compared to 8 motors for the PD and torque-controllers. Considering that the training time is mostly below an hour on a consumer-GPU, this computational efficiency is sufficient for the purpose of this study. The environment runs with a timestep of $5$~ms for the physics engine and a control time step of $20$~ms.

For the hardware experiments, the muscle model is implemented in C++, which communicates with real-time Python functions which were bound via PyBind11~\cite{pybind11}. This module interfaces with the RL policy running at $\sim 50$~Hz, which is not real-time executed, while the actual muscle-controller is running at $\sim 500$~Hz. It is important to achieve a large execution frequency on the hardware to reliably emulate the muscle model~\cite{Wochneretal22}. This paradigm is similar to most studies using PD-controllers on quadruped robots~\cite{torque_quadruped}.

\subsection{Muscle parameter tuning}
In contrast to the PD-controller~\cite{ziegler_pd}, there is no clear tuning procedure for a muscle-controller. The velocity scaling $\beta$ is especially difficult to tune, as it affects the damping of the actuator through the FV-relationship~\cite{Wochneretal22} and might not transfer to real hardware due to control frequency limitations, see \secref{sec:params}. When we tried to set this parameter through an iterative tuning procedure similar to previous studies~\cite{mattern2023mimo}, we either obtained parameters that performed well in simulation but were not stable on the hardware, or they were stable but did not yield performant policies. 

We derive an expression for the damping coefficient $\beta$, such that under certain conditions, the damping is equivalent to a damping-controller $\tau = -k_{\mathrm{damp}}\dot{q}$. With this approach, we can determine reasonable $\beta$ values that are achievable on the hardware based on the easier-to-tune damping-controller. 

We start by assuming two symmetric muscles connected to a single joint at position $q=0$. As the FV-relationship is thought to primarily contribute to muscle damping~\cite{Izzi2023}, we also assume $\mathrm{FL}(l)=1$, $\mathrm{FP}(l)=0$ and constant activity $m_{act,k}=1$. \Eqnref{eq:muscle} then simplifies to: 
\begin{align}
    \tau &= f_{\mathrm{max}} \, \left(\mathrm{FV}(\bar{\dot{l}}_{1}) - \mathrm{FV}(\bar{\dot{l}}_{2})\right)\\
       &= f_{\mathrm{max}} \, \left(\mathrm{FV}(\bar{\dot{l}}_{1}) - \mathrm{FV}(-\bar{\dot{l}}_{1})\right)\\
       &\approx f_{\mathrm{max}} \,(4\,\bar{\dot{l}}_{1})
       \label{eq:muscle_taylor}
\end{align}
where $\bar{\dot{l}}_{1}=-\bar{\dot{l}}_{2}$ in \eqnref{eq:muscle_vel} as we assume symmetric muscles and $a_{1}=-a_{2}$ in \eqnref{eq:muscle_len}, and we have used a Taylor expansion around $\bar{\dot{l}}_{1}=0$. By using the definition of the FV-curve from~\cite{Wochneretal22, mujoco}, we obtain \[\mathrm{FV}(\bar{\dot{l}})=1+2\,\bar{\dot{l}} +\mathcal{O}(\bar{\dot{l}}^{2}).\]
By comparing \eqnref{eq:muscle_taylor} to a damping controller and using $\bar{\dot{l}}_{1} = a_{1} \beta \dot{q}$ we get:
\begin{align}
    - k_{\mathrm{damp}} \dot{q} &= - f_{\mathrm{max}}\,4\,\beta\,a_{1}\,\dot{q} \label{eq:neg_sign} \\
    \Rightarrow \beta &= \frac{k_{\mathrm{damp}}}{4\,a_{1}f_{\mathrm{max}}}
    \label{eq:damping_rule}.
\end{align}
The negative sign in \eqnref{eq:neg_sign} was introduced as muscles are assumed to always pull on the joint. We found that setting $k_{\mathrm{damp}} =0.1$ is an \textit{achievable} value for a damping controller on the robot hardware. By using \eqnref{eq:damping_rule} for all muscle experiments, we ensure that the maximum co-contraction level generates \textit{achievable} damping on the hardware. The policy can still reduce the amount of equivalent damping by changing the amount of co-contraction~\cite{Izzi2023}. 

\subsection{Tasks}
We employ two tasks: walking and hopping, which have been proven interesting in previous comparisons of action spaces~\cite{Wochneretal22}. Walking requires a periodic and relatively slow joint movement, while periodic hopping is a highly explosive movement that also requires stabilization when landing.

\subsubsection{Walking}
We restrict ourselves to the target velocity tracking reward from~\cite{rudin2021learning} and omit all other terms that were used in that study:
\begin{equation}
    r_{\mathrm{walk}} = \mathrm{exp}\left(- (v_{\mathrm{target}} - v_{x})^{2}/\sigma\right),
\end{equation}
where $v_{\mathrm{target}}$ is the target x-velocity, $v_{x}$ is the x-velocity of the base and $\sigma = 0.25$ is a sensitivity factor. The task is reset if the robot base or the upper legs make contact with the ground or if the base height is $h_{\mathrm{base}} < 0.2$\,m.

\subsubsection{Periodic hopping}
The hopping reward is based on the vertical velocity of the robot base $v_{z}$ as 
\begin{equation}
    r_{\mathrm{hop}} = \mathrm{exp}\left(10\,v_{\mathrm{clip}}\right),
\label{eq:hop}
\end{equation}
where $v_{\mathrm{clip}} = \mathrm{clip}(v_{z}, 0, 10)$. This is the same hopping reward as in~\cite{Wochneretal22} with slightly different scaling, accounting for a different velocity range with the SOLO robot. We observed the reward function to not work well if the desired hopping velocity is in a range where the reward function is not sensitive to changes. The episode is reset when the robot base changes in orientation too much from upright orientation, or when a body part different than the feet touches the ground.

Both tasks additionally use action rate regularization:
\begin{equation}
r_{\mathrm{act}} = - w_{\mathrm{act}}\,\left(a_{t+1} - a_{t}\right)^{2},
\label{eq:action_rate}
\end{equation}
where $a_{t}$ is the action at time step $t$ and $w_{\mathrm{act}}$ is a weighting coefficient.

The total reward is
\begin{equation}
    r=r_\mathrm{walk/hop}+r_\mathrm{act}.
\end{equation}

\subsection{Learning high-level policies}
In order to train policies, we use the popular RL algorithm PPO~\cite{schulman2017proximal} with the implementation from~\cite{rudin2021learning} and identical hyperparameters as in~\cite{li2022wasabi} for the \texttt{SoloLeap}-task, shown in Table S6 in their work. Our policy and critic are fully-connected MLPs with 3 hidden dimensions of 128 units each and ELU-activation functions. For simulation experiments, we do not use domain randomization, while sim-to-real transfer requires randomization of initial states, ground friction, body mass, and input noise. See \tabref{tab:DR} for details.
\begin{table}[ht]
 \caption{Values for domain randomization~(DR) and input noise~(IN). All values are sampled uniformly in the given ranges and added to the existing values.}
    \label{tab:DR}
    \centering
    \begin{tabular}{@{}lll@{}}
        \toprule
        &\textbf{Parameter} & \textbf{Range} \\

        \midrule
           \multirow{6}{*}{DR} & init. joint pos. & $[-1.0, 1.0]$ \\
           & init. muscle act. & $[0.5, 1.0]$ \\
           & friction & $[0.5, 1.25]$ \\
           & joint damping & $[0, 0.09]$\\
           & push$_\mathrm{xy}$ & $[-1.5, 1.5]$ \\
           & mass shift & $[-0.5, 1.2]$ \\
         \midrule
          \multirow{9}{*}{IN} & base lin. vel.  & $[-0.02, 0.02]$ \\
           & base ang. vel.  & $[-0.05, 0.05]$ \\
           & gravity vector  & $[-0.05, 0.05]$ \\
           & joint pos.  & $[-0.01, 0.01]$ \\
           & joint vel.  & $[-0.075, 0.075]$ \\
           & muscle length  &$[-0.01, 0.01]$ \\
           & muscle vel.  & $[-1.0, 1.0]$ \\
           & muscle act.  & $[-0.01, 0.01]$ \\
           & muscle force  & $[-1.0, 1.0]$ \\
        \bottomrule
        \end{tabular}
\end{table}

\section{Results}
Each simulation experiment was repeated over 10 random seeds. In experiments where we show a specific rollout of a specific policy, we analyzed all trained seeds for all policies at the end of their training, and selected in each case the one with the best behavior. In general we observe very small variance across seeds, likely due to the large number~(4096) of parallel environments in Isaac Gym.\looseness-1

\subsection{Parameter tuning procedure}
\label{sec:params}
\begin{table}[h]
    \caption{Optimized parameter values for the used actuators. Note that $\beta$ is not optimized but set through \eqnref{eq:damping_rule}.}
    \label{tab:actuator_parameters}
    \centering
    \begin{tabular}{@{}lll@{}}
        \toprule
        &\textbf{Parameter} & \textbf{Value} \\
        \midrule
         \multirow{11}{*}{muscle} & $l_{\mathrm{min}}$ & 0.24 \\
         & $l_{\mathrm{max}}$ & 1.53\\
         & $\mathrm{fvmax}$ & 1.38 \\
         & $\mathrm{fpmax}$ & 1.76 \\
         & $\mathrm{lce\_min}$ & 0.74 \\
         & $\mathrm{lce\_max}$ & 0.94 \\
         & $f_\mathrm{max}$ & 34 \\
         & $\phi_{\mathrm{min}}$ & -3.14 \\
         & $\phi_{\mathrm{max}}$ & 3.14\\
         & $\tau_{\mathrm{act}}$ & $0.01$\\
         & {\color{ourred}$\beta$} & {\color{ourred}$0.36$}\\

         \midrule
         \multirow{2}{*}{PD} & $\stiffness$ & $2$ \\
         & $\damping$  & $0.05$ \\
        \midrule
        torque & n.a. & n.a. \\
        \midrule
        action rate weight & $ w_{\mathrm{act}}$ & 0.004\\
        \bottomrule
        \end{tabular}
\end{table}

As our study aims to investigate the potential benefits of muscle-like actuators, we use a methodical approach for parameter tuning. First, only the task-relevant reward is used as objective. We tune the actuator-relevant parameters to achieve maximum performance in a population-based optimization over 10 iterations of 100 trials each. This optimization successfully leads to policies achieving large task rewards. Upon closer inspection, however, we observe them to be very jittery, not likely to generalize to the real hardware. We consequently add action-rate regularization, see \eqnref{eq:action_rate}.
The best parameters from the previous step are taken, and a grid search over weighting coefficients for the action rate cost is performed. It is commonly observed that smooth actuation signals are more transferable to real-world robots~\cite{mysore2021regularizing}. The level of action smoothness that works well for sim-to-real transfer is hard to quantify, as such we visually check for the variance in the applied torques, instead of optimizing the regularizer together with the other parameters.  

This optimization procedure is executed for the PD and muscle low-level controller for the walking task. The torque actuator has no parameters besides the maximum torque, which is given by the system specification of the robot. We then keep all parameters identical for the hopping task.
\begin{figure}[htbp]
\centering
\hspace{20pt} \textcolor{ourorange2}{\rule[2.5pt]{15pt}{1.5pt}} $\beta=0.36$ \,\,\textcolor{ourpurple3}{\rule[2.5pt]{15pt}{1.5pt}} $\beta=0.66$\\
\vspace{6pt}
\includegraphics[height=0.25\textheight]{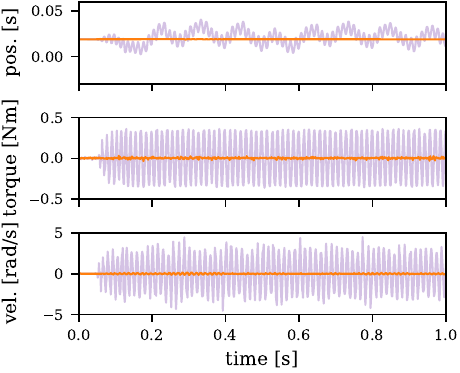}
\caption{\textbf{Hardware deployment: excessive emulated damping destabilizes the control.} We apply constant muscle excitations of $m_{\mathrm{act}}=1.0$ on both muscles of a joint of the hardware robot. Ideally, the joint would remain at its initial position, resisting external perturbations with high damping and stiffness. When using a high scaling value $\beta=0.66$, the emulation controller can not keep up due to its limited control frequency, and the motor starts to oscillate strongly. The value $\beta=0.36$, given by \eqnref{eq:damping_rule}, leads to stable performance. We observe similar effects with a PD-controller and excessive $k_{\mathrm{damp}}$-values.}
\label{fig:damping}
\end{figure}

\subsection{Walking}
While performant and transferable gaits usually require dozens of cost terms which are hand-tuned by experts~\cite{li2022wasabi, rudin2021learning}, we investigate the behaviors that arise from a simple and under-specified forward velocity reward combined with an action rate cost. Different actuator morphologies might embed certain movement priors into the learning algorithm which do not require a complex tuning procedure.

\begin{figure}[htbp]
\centering
\hspace{20pt}\textcolor{ourblue2}{\rule[2.5pt]{15pt}{1.5pt}} PD \,\,\textcolor{ourorange2}{\rule[2.5pt]{15pt}{1.5pt}} muscle \,\,\textcolor{ourgreen2}{\rule[2.5pt]{15pt}{1.5pt}} torque\\
\vspace{6pt}
\includegraphics[height=0.25\textheight]{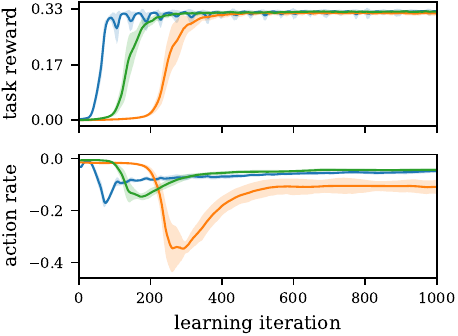}
\caption{\textbf{Stable locomotion is achieved by all actuators.} Upper: Velocity-tracking task reward over time. The PD-agent learns with the smallest number of iterations, while torque- and muscle-agents achieve a similar maximum performance with more training. Lower: The action rate reward only slowly improves for the muscle-agent, likely because the number of actions is twice as large as for the other actuators (16 vs 8).}
\label{fig:performance_walking}
\end{figure}

\begin{figure*}[htbp]
    \centering
      \begin{subfigure}[b]{0.49\textwidth}
      \hspace{2.6cm} \textcolor{ourblue2}{\rule[2.5pt]{15pt}{1.5pt}} PD \,\,\textcolor{ourorange2}{\rule[2.5pt]{15pt}{1.5pt}} muscle\,\,\textcolor{ourgreen2}{\rule[2.5pt]{15pt}{1.5pt}} torque\\
      \\
      \includegraphics[width=1.05\linewidth]{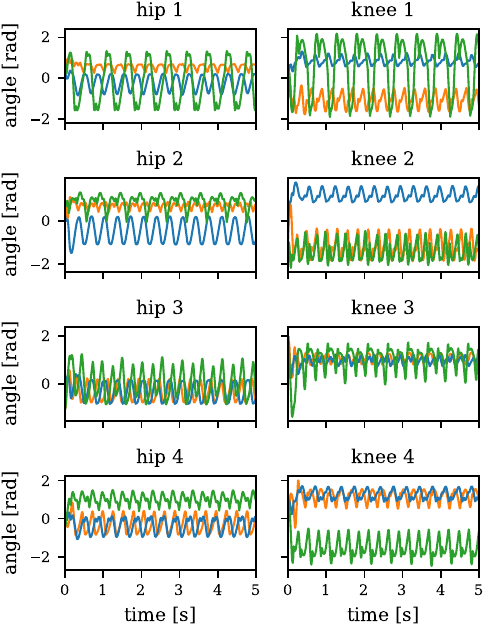}
      \caption{Joint angles over time.}
    \end{subfigure}
    \begin{subfigure}[b]{0.45\textwidth}
      \centering
      \hspace{0.4cm}
      \includegraphics[width=0.93\linewidth]{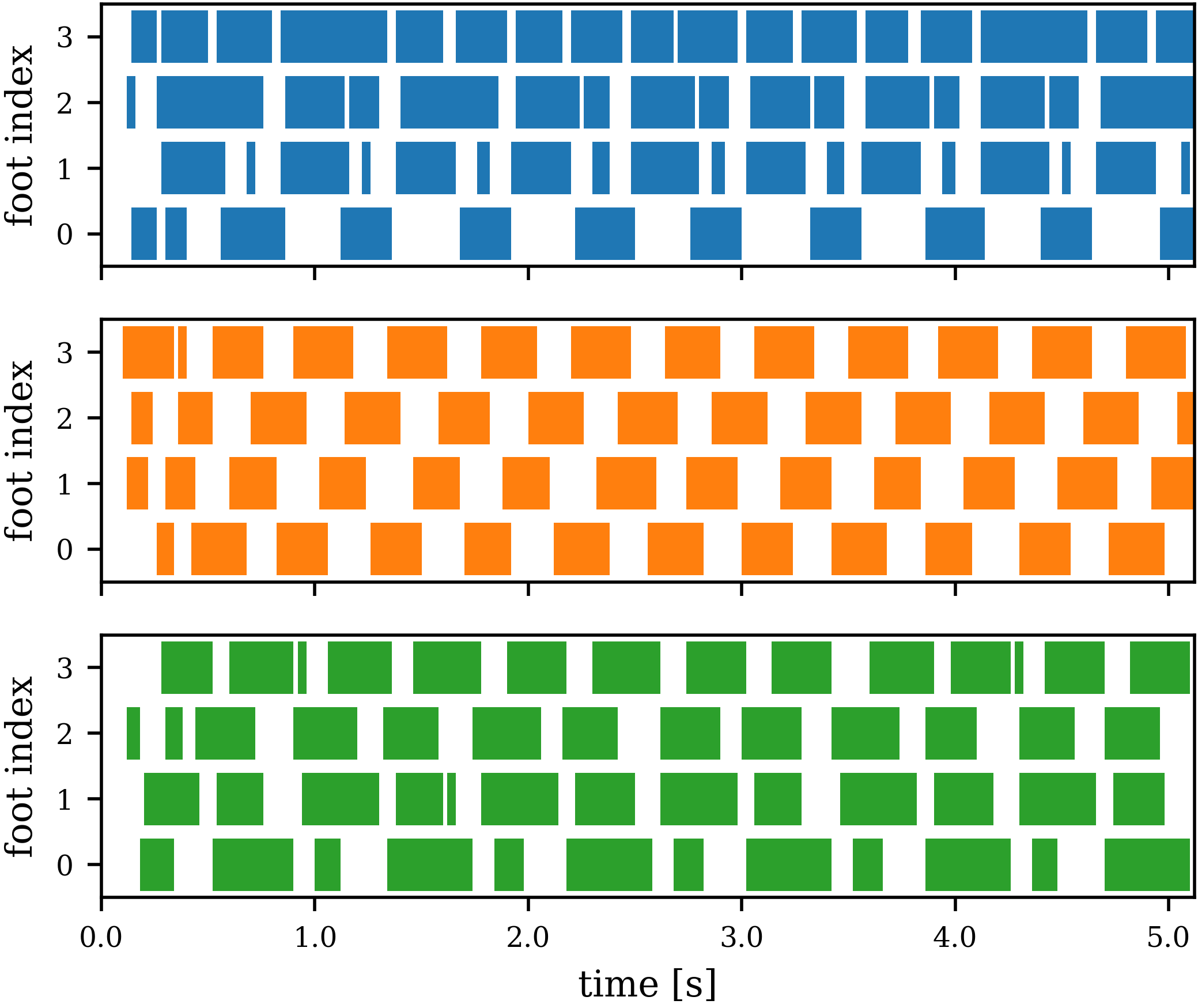}\\
      \caption{Foot-contact patterns.}
      \vspace{0.1cm}
      \begin{subfigure}{1.0\textwidth}
        {\hspace{0.65cm}\includegraphics[width=0.92\linewidth]{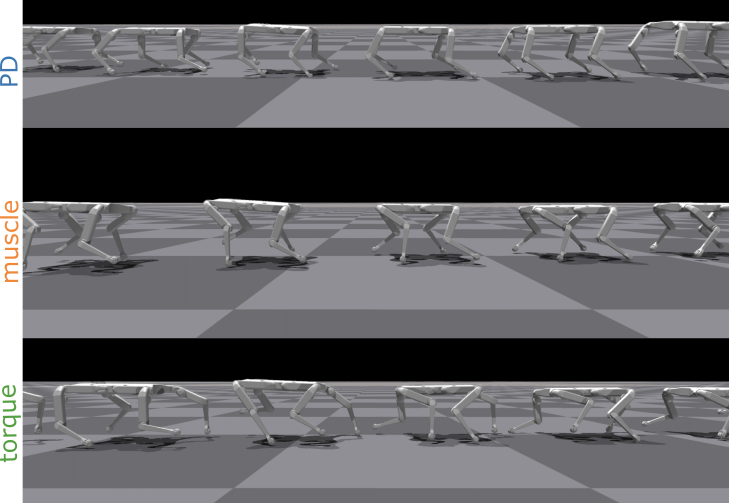}\\
        \caption{Learned gaits.}}
      \end{subfigure}
    \end{subfigure}\\
    \caption{\textbf{Muscles learn more regular walking patterns.} Using only the linear velocity and an action rate regularizing term as reward, the muscle-actuator learns the most regular walking pattern, lifting the feet above the ground while moving. In contrast, the torque-actuator agent walks with strongly extended motions that come close to the angular limits of the robotic limbs. The PD-controller agent drags the feet very closely over the ground, likely not generalizing to real-world hardware.}
    \label{fig:walking_gaits}
\end{figure*}

The task reward over training improves faster for the PD- and torque-controllers than for the muscle-controllers, see \figref{fig:performance_walking}. While all three policies perform well according to this metric, we also take a look at the learned gaits in \figref{fig:walking_gaits}.

The touch patterns of the feet with the ground (\figref{fig:walking_gaits}b) show that the muscle-agent learns the most regular gait. The PD-agent tends to not elevate the feet from the ground, as can be seen from the gait illustration (\figref{fig:walking_gaits}c) and the joint angle trajectories (\figref{fig:walking_gaits}a). This behavior relies on very precise movements and friction coefficients in the simulation, which would likely not generalize to the real world. Interestingly, the torque-agent lifts the feet off the ground but tends to use extreme and uncontrolled motions. This showcases that even with an under-specified reward function, a particular embodiment, or low-level controller, can bias the behavior towards being more natural.

\subsection{Periodic hopping}
\begin{figure*}[htbp]
\centering
\begin{subfigure}{0.65\textwidth}
\centering
  \textcolor{ourblue2}{\rule[2.5pt]{15pt}{1.5pt}} PD \,\, \hspace{10pt}\textcolor{ourorange2}{\rule[2.5pt]{15pt}{1.5pt}} muscle\hspace{10pt}\textcolor{ourgreen2} {\rule[2.5pt]{15pt}{1.5pt}} torque\\
          \vspace{6pt}
\begin{subfigure}{1.0\textwidth}

  {\hspace{0.58cm}\includegraphics[width=0.9\linewidth]{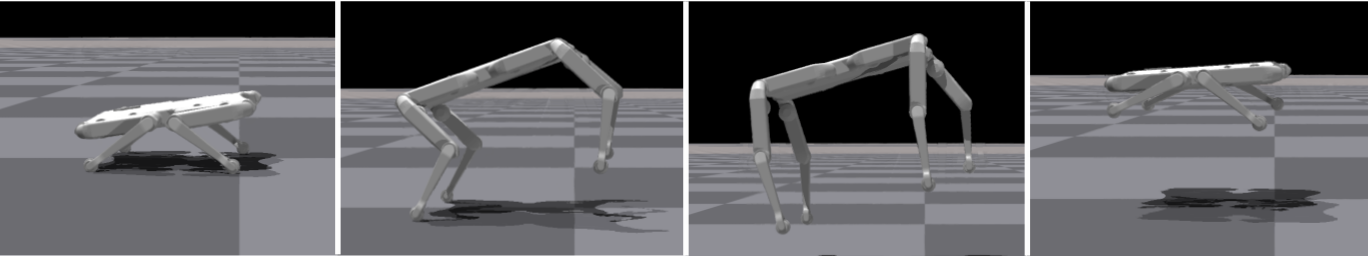}}
  \caption{Hopping motion.}
  \end{subfigure}\\
  \centering
{\includegraphics[height=0.3\linewidth]{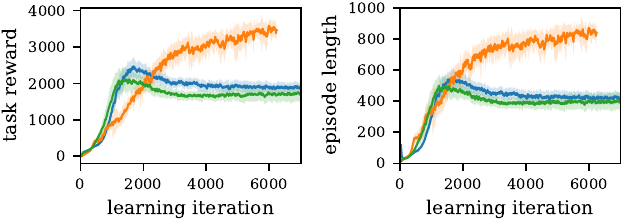}}
\caption{Task performance.}
\end{subfigure}
\begin{subfigure}{0.3\textwidth}
\centering
{\hspace{-9pt}\includegraphics[height=1.25\linewidth]{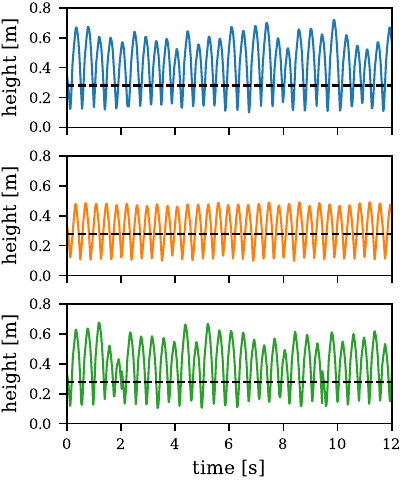}}
\caption{Height of the robot base.}
\end{subfigure}
\caption{\textbf{Muscles learn more stable hopping.} Similar to results by Wochner et al.~\cite{Wochneretal22}, we observe that muscle actuators learn more stable periodic hopping than torque or PD actuators. While the torque and especially the PD agents can occasionally achieve larger maximum heights than the muscle-agent, the periodic hopping behavior is less stable and achieves a smaller overall episode return. This task rewards large vertical velocities exponentially, the optimal behavior is therefore periodic jumping with maximum velocity and large displacement. We show the average base height while walking as a black line in (c) as a reference height for the hopping motion.}  
\label{fig:performance_hopping}
\end{figure*}

The simulation results for the maximum height periodic hopping task are shown in \figref{fig:performance_hopping}. The reward function, see \eqnref{eq:hop}, incentivizes maximal upwards velocities for the longest possible time. Even though the PD and torque agents achieve some very high jumps (\figref{fig:performance_hopping}c), the hopping behavior is straighter and more regular with the muscle actuation, making it easier to control when landing. This leads to large episode lengths (\figref{fig:performance_hopping}b) for the muscle-agent, as the other controllers can often not recover after landing. 
These results are slightly different to \cite{Wochneretal22}, where the muscle-agent achieved better performance than the alternatives only early in the training, but this difference may be explained by the superior stability of quadrupeds over bipedal robots.

\subsection{Robustness}
\begin{figure*}[htbp]
\begin{subfigure}{0.49\linewidth}
\centering
\includegraphics[height=0.15\textheight]{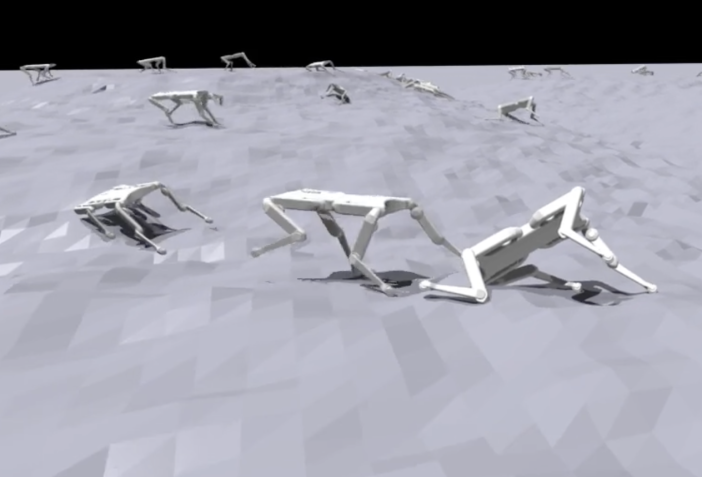}
\caption{Difficult terrain.}
\end{subfigure}
\begin{subfigure}{0.49\linewidth}
\centering
\includegraphics[height=0.15\textheight]{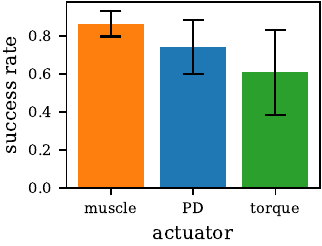}
\caption{Success rates for different actuators.}
\end{subfigure}
\caption{\textbf{The muscle actuator is the most robust.} We rollout the different agents in an unseen terrain. The success rate is highest for the muscle actuator, followed by the PD and then the torque.}
\label{fig:robustness_plot}
\end{figure*}

Wochner et al.~\cite{Wochneretal22} demonstrated that muscle-actuators exhibit higher robustness under unseen perturbations. We therefore tested the trained policies under terrain variations used in~\cite{rudin2021learning} and recorded the fraction of steps across 100 episodes in which the policies would not fall down, which we define as the success rate. Note that the policies have only been trained on flat terrain.
We see that the muscle-controller is generally more robust and has higher success rates than the other controllers (\figref{fig:robustness_plot}).

\subsection{Sim-to-real transfer}
\begin{figure*}[htbp]
\centering
\hspace{30pt}\textcolor{ourorange2}{\rule[2.5pt]{15pt}{1.5pt}} muscle \,\,\textcolor{ourblue2}{\rule[2.5pt]{15pt}{1.5pt}} PD\\
          \vspace{6pt}
\begin{subfigure}{1.0\linewidth}
\centering
\includegraphics[height=0.23\textheight]{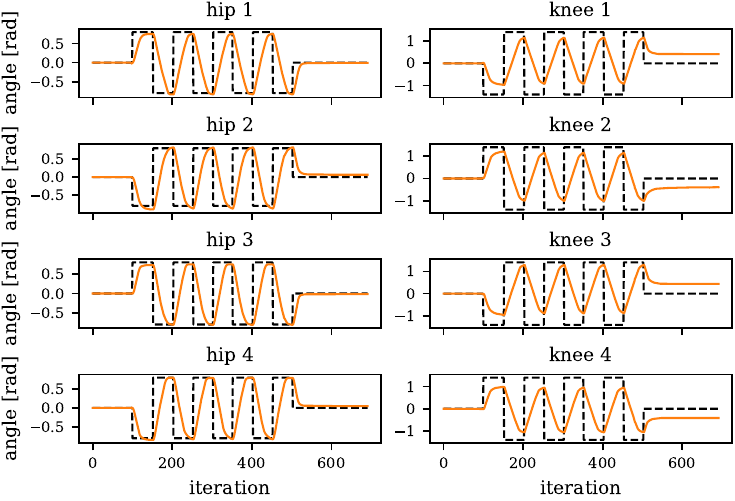}
\includegraphics[height=0.23\textheight]{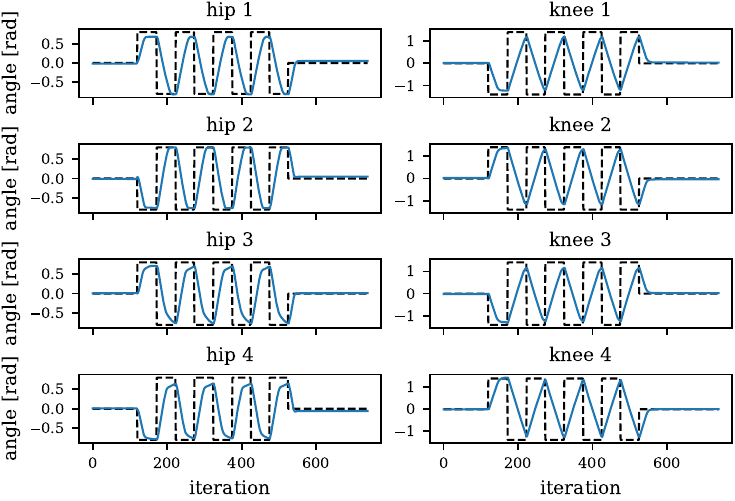}
\caption{Joint angle tracking with muscle and PD actuators.}
\end{subfigure}\\
\vspace{5pt}
\begin{subfigure}{0.9\linewidth}
\centering
\includegraphics[height=0.13\textheight]{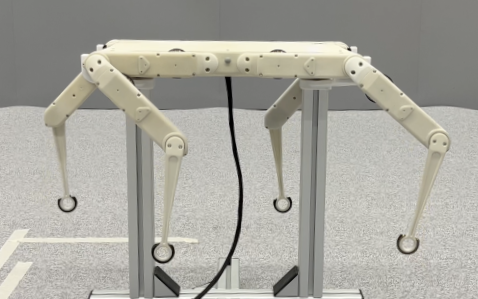}
\includegraphics[height=0.13\textheight]{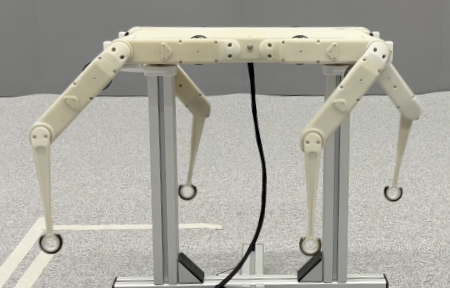}
\caption{Exemplary target angles on the real robot.}
\end{subfigure}
\begin{subfigure}{1.0\linewidth}
\centering
\includegraphics[height=0.15\textheight]{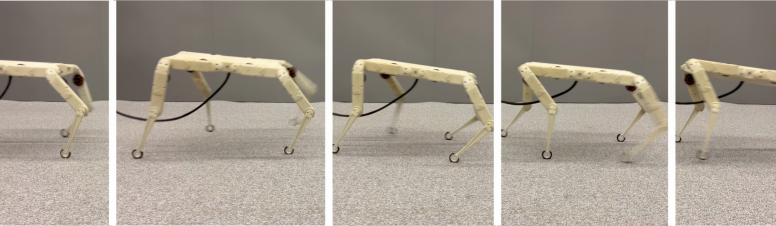}
\caption{Walking gait of the muscle-policy.}
\label{fig:hardware_gait}
\end{subfigure}

\caption{\textbf{Sim-to-real transfer with an emulated muscle actuator.} (a) \& (b) We trained policies for joint angle target tracking in simulation under the addition of simulated sensor noise and domain randomization. The muscle-policy is able to track targets with a real-time emulated muscle actuator on real robotic hardware through sim2real transfer. Note that the target angles were given at a fast frequency, making it challenging to track them perfectly in order to achieve a dynamic regime. (c) We train a muscle-driven policy in a walking task in simulation and deploy it on the real robot hardware.}
\label{fig:performance_hardware}
\end{figure*}

In order to test the applicability of our findings and highlight their relevance for future robot design, we implemented our muscle model on real robotic hardware. First, we demonstrate the sim-to-real transfer of our policies in a task requiring tracking joint angles over time. For this experiment, the robot is suspended in mid-air on a fixed stand (\figref{fig:performance_hardware}b).

The sim-to-real policies were trained with observation noise and domain randomization, see \tabref{tab:DR}. Finally, we added a small amount of joint damping through a low-level controller with $k_{\mathrm{damp}}=0.08$ Nms/rad in all hardware experiments to stabilize the system and prevent damage. We also added it in simulation to minimize the sim-to-real gap. 

We compare a policy trained with the real-time muscle-controller to a PD-controller for which we give alternating target positions and set the target velocity to zero (\figref{fig:performance_hardware}a). We use PD gains $\stiffness=5.0$ and $\damping=0.1$, similar to~\cite{li2022wasabi}. \Figref{fig:performance_hardware}a shows that the muscle-controller is able to track the given joint angles similarly well to the PD-controller.

In a more realistic task, we train a walking policy with the muscle-controller in simulation and can reliably perform a running gait without additional training on the hardware. This successful sim-to-real transfer demonstrates that muscle actuation enables learning realistic and robust gaits without the need for excessive reward engineering. See \figref{fig:hardware_gait} and the videos on the project website \textcolor{blue}{\url{https://sites.google.com/view/emulatedmuscles}}.

\section{Conclusion}
In this study, we have investigated the benefits of muscle-like actuation for quadruped locomotion tasks. We validated previous results on stability and robustness of the learned policies, while finding that the emulated actuator morphology also influences the learned gaits in an under-specified locomotion task. Finally, we proposed a tuning procedure that allowed us to validate the muscle model on a real robotic system under performance and stability constraints. We are able to execute policies learned entirely in simulation with little reward engineering on robotic hardware that runs a real-time emulated muscle model. These results showcase the potential of muscle-like actuators when combined with RL.

\bibliographystyle{IEEEtran}
\bibliography{IEEEabrv,literature}

\begin{thebibliography}{10}
\providecommand{\url}[1]{#1}
\csname url@samestyle\endcsname
\providecommand{\newblock}{\relax}
\providecommand{\bibinfo}[2]{#2}
\providecommand{\BIBentrySTDinterwordspacing}{\spaceskip=0pt\relax}
\providecommand{\BIBentryALTinterwordstretchfactor}{4}
\providecommand{\BIBentryALTinterwordspacing}{\spaceskip=\fontdimen2\font plus
\BIBentryALTinterwordstretchfactor\fontdimen3\font minus \fontdimen4\font\relax}
\providecommand{\BIBforeignlanguage}[2]{{%
\expandafter\ifx\csname l@#1\endcsname\relax
\typeout{** WARNING: IEEEtran.bst: No hyphenation pattern has been}%
\typeout{** loaded for the language `#1'. Using the pattern for}%
\typeout{** the default language instead.}%
\else
\language=\csname l@#1\endcsname
\fi
#2}}
\providecommand{\BIBdecl}{\relax}
\BIBdecl

\bibitem{2017-SCA-action}
\BIBentryALTinterwordspacing
X.~B. Peng and M.~van~de Panne, ``Learning locomotion skills using deeprl: Does the choice of action space matter?'' in \emph{Proceedings of the ACM SIGGRAPH / Eurographics Symposium on Computer Animation}, ser. SCA '17.\hskip 1em plus 0.5em minus 0.4em\relax New York, NY, USA: ACM, 2017, pp. 12:1--12:13. [Online]. Available: \url{http://doi.acm.org/10.1145/3099564.3099567}
\BIBentrySTDinterwordspacing

\bibitem{bohg2019}
R.~Martín-Martín, M.~A. Lee, R.~Gardner, S.~Savarese, J.~Bohg, and A.~Garg, ``Variable impedance control in end-effector space: An action space for reinforcement learning in contact-rich tasks,'' in \emph{2019 IEEE/RSJ International Conference on Intelligent Robots and Systems (IROS)}, 2019, pp. 1010--1017.

\bibitem{aljalbout2023role}
E.~Aljalbout, F.~Frank, M.~Karl, and P.~van~der Smagt, ``On the role of the action space in robot manipulation learning and sim-to-real transfer,'' 2023.

\bibitem{Wochneretal22}
\BIBentryALTinterwordspacing
I.~Wochner, P.~Schumacher, G.~Martius, D.~B{\"u}chler, S.~Schmitt, and D.~Haeufle, ``Learning with muscles: Benefits for data-efficiency and robustness in anthropomorphic tasks,'' in \emph{Proceedings of the 6th Conference on Robot Learning (CoRL)}, ser. Proceedings of Machine Learning Research, vol. 205.\hskip 1em plus 0.5em minus 0.4em\relax PMLR, Dec. 2022, pp. 1178--1188. [Online]. Available: \url{https://proceedings.mlr.press/v205/wochner23a.html}
\BIBentrySTDinterwordspacing

\bibitem{vnmc2015}
Z.~Batts, S.~Song, and H.~Geyer, ``Toward a virtual neuromuscular control for robust walking in bipedal robots,'' in \emph{2015 IEEE/RSJ International Conference on Intelligent Robots and Systems (IROS)}, 2015, pp. 6318--6323.

\bibitem{atras2018}
A.~Rai, R.~Antonova, S.~Song, W.~Martin, H.~Geyer, and C.~Atkeson, ``Bayesian optimization using domain knowledge on the atrias biped,'' in \emph{2018 IEEE International Conference on Robotics and Automation (ICRA)}, 2018, pp. 1771--1778.

\bibitem{vansoest}
A.~J. van Soest and M.~F. Bobbert, ``{{T}he contribution of muscle properties in the control of explosive movements},'' \emph{Biol Cybern}, vol.~69, no.~3, pp. 195--204, 1993.

\bibitem{Izzi2023}
\BIBentryALTinterwordspacing
F.~Izzi, A.~Mo, S.~Schmitt, A.~Badri-Spr{\"o}witz, and D.~F.~B. Haeufle, ``Muscle prestimulation tunes velocity preflex in simulated perturbed hopping,'' \emph{Scientific Reports}, vol.~13, no.~1, p. 4559, Mar 2023. [Online]. Available: \url{https://doi.org/10.1038/s41598-023-31179-6}
\BIBentrySTDinterwordspacing

\bibitem{Araz2023}
M.~Araz, S.~Weidner, F.~Izzi, A.~Badri-Spr{\"o}witz, T.~Siebert, and D.~F.~B. Haeufle, ``\BIBforeignlanguage{en}{Muscle preflex response to perturbations in locomotion: In vitro experiments and simulations with realistic boundary conditions},'' \emph{\BIBforeignlanguage{en}{Front Bioeng Biotechnol}}, vol.~11, p. 1150170, Apr. 2023.

\bibitem{Haeufle_2010}
\BIBentryALTinterwordspacing
D.~F.~B. Haeufle, S.~Grimmer, and A.~Seyfarth, ``The role of intrinsic muscle properties for stable hopping---stability is achieved by the force--velocity relation,'' \emph{Bioinspiration \& Biomimetics}, vol.~5, no.~1, p. 016004, feb 2010. [Online]. Available: \url{https://dx.doi.org/10.1088/1748-3182/5/1/016004}
\BIBentrySTDinterwordspacing

\bibitem{Seyfarth2007}
\BIBentryALTinterwordspacing
A.~Seyfarth, K.~T. Kalveram, and .~Geyer, Hartmut, ``{Simulating Muscle-Reflex Dynamics in a Simple Hopping Robot},'' in \emph{Proceedings of Fachgespr{\"{a}}che Autonome Mobile Systeme}.\hskip 1em plus 0.5em minus 0.4em\relax Springer, 2007, pp. 294--300. [Online]. Available: \url{http://link.springer.com/10.1007/978-3-540-74764-2_45}
\BIBentrySTDinterwordspacing

\bibitem{Mo_2023}
A.~Mo, F.~Izzi, E.~C. Gönen, D.~Haeufle, and A.~Badri-Spröwitz, ``Slack-based tunable damping leads to a trade-off between robustness and efficiency in legged locomotion,'' \emph{Scientific Reports}, vol.~13, no.~1, feb 2023.

\bibitem{mckibben2006}
B.~Tondu and S.~Zagal, ``Mckibben artificial muscle can be in accordance with the hill skeletal muscle model,'' in \emph{The First IEEE/RAS-EMBS International Conference on Biomedical Robotics and Biomechatronics, 2006. BioRob 2006.}, 2006, pp. 714--720.

\bibitem{Ma2023}
\BIBentryALTinterwordspacing
H.~Ma, D.~B{\"u}chler, B.~Sch{\"o}lkopf, and M.~Muehlebach, ``Reinforcement learning with model-based feedforward inputs for robotic table tennis,'' \emph{Autonomous Robots}, vol.~47, no.~8, pp. 1387--1403, Dec 2023. [Online]. Available: \url{https://doi.org/10.1007/s10514-023-10140-6}
\BIBentrySTDinterwordspacing

\bibitem{hasel2021}
\BIBentryALTinterwordspacing
P.~Rothemund, N.~Kellaris, S.~K. Mitchell, E.~Acome, and C.~Keplinger, ``Hasel artificial muscles for a new generation of lifelike robots—recent progress and future opportunities,'' \emph{Advanced Materials}, vol.~33, no.~19, p. 2003375, 2021. [Online]. Available: \url{https://onlinelibrary.wiley.com/doi/abs/10.1002/adma.202003375}
\BIBentrySTDinterwordspacing

\bibitem{buchler2022learning}
D.~B{\"u}chler, S.~Guist, R.~Calandra, V.~Berenz, B.~Sch{\"o}lkopf, and J.~Peters, ``Learning to play table tennis from scratch using muscular robots,'' \emph{IEEE Transactions on Robotics}, vol.~38, no.~6, pp. 3850--3860, 2022.

\bibitem{Guistetal23}
\BIBentryALTinterwordspacing
S.~Guist, J.~Schneider, A.~Dittrich, V.~Berenz, B.~Sch{\"o}lkopf, and D.~B{\"u}chler, ``Hindsight states: Blending sim and real task elements for efficient reinforcement learning,'' in \emph{Robotics: Science and Systems XIX}, Jul. 2023. [Online]. Available: \url{https://www.roboticsproceedings.org/rss19/p038.html}
\BIBentrySTDinterwordspacing

\bibitem{dcmuscle2006}
H.~Serhan, C.~Nasr, and P.~Henaff, ``Designing a muscle like system based on pid controller and tuned by neural network,'' in \emph{The 2006 IEEE International Joint Conference on Neural Network Proceedings}, 2006, pp. 4991--4998.

\bibitem{dcmuscle_knee2016}
\BIBentryALTinterwordspacing
H.~Serhan and P.~Henaff, ``Muscle‐like compliance in knee articulations improves biped robot walkings,'' in \emph{Recent Advances in Robotic Systems}, G.~Wang, Ed.\hskip 1em plus 0.5em minus 0.4em\relax Rijeka: IntechOpen, 2016, ch.~3. [Online]. Available: \url{https://doi.org/10.5772/63746}
\BIBentrySTDinterwordspacing

\bibitem{schneider2023investigating}
J.~Schneider, P.~Schumacher, D.~Häufle, B.~Schölkopf, and D.~Büchler, ``Investigating the impact of action representations in policy gradient algorithms,'' 2023.

\bibitem{makoviychuk2021isaac}
V.~Makoviychuk, L.~Wawrzyniak, Y.~Guo, M.~Lu, K.~Storey, M.~Macklin, D.~Hoeller, N.~Rudin, A.~Allshire, A.~Handa, and G.~State, ``Isaac gym: High performance gpu-based physics simulation for robot learning,'' 2021.

\bibitem{rudin2021learning}
\BIBentryALTinterwordspacing
N.~Rudin, D.~Hoeller, P.~Reist, and M.~Hutter, ``Learning to walk in minutes using massively parallel deep reinforcement learning,'' in \emph{5th Annual Conference on Robot Learning}, 2021. [Online]. Available: \url{https://openreview.net/forum?id=wK2fDDJ5VcF}
\BIBentrySTDinterwordspacing

\bibitem{li2022wasabi}
\BIBentryALTinterwordspacing
C.~Li, M.~Vlastelica, S.~Blaes, J.~Frey, F.~Grimminger, and G.~Martius, ``Learning agile skills via adversarial imitation of rough partial demonstrations,'' in \emph{Proceedings of the 6th Conference on Robot Learning (CoRL)}, Dec. 2022. [Online]. Available: \url{https://openreview.net/forum?id=x6INXlnUGro}
\BIBentrySTDinterwordspacing

\bibitem{torque_quadruped}
S.~Chen, B.~Zhang, M.~W. Mueller, A.~Rai, and K.~Sreenath, ``Learning torque control for quadrupedal locomotion,'' in \emph{2023 IEEE-RAS 22nd International Conference on Humanoid Robots (Humanoids)}, 2023, pp. 1--8.

\bibitem{pybind11}
W.~Jakob, J.~Rhinelander, and D.~Moldovan, ``pybind11 -- seamless operability between {C++11} and {Python},'' 2017, https://github.com/pybind/pybind11.

\bibitem{ziegler_pd}
\BIBentryALTinterwordspacing
J.~G. Ziegler and N.~B. Nichols, ``{Optimum Settings for Automatic Controllers},'' \emph{Transactions of the American Society of Mechanical Engineers}, vol.~64, no.~8, pp. 759--765, 12 2022. [Online]. Available: \url{https://doi.org/10.1115/1.4019264}
\BIBentrySTDinterwordspacing

\bibitem{mattern2023mimo}
D.~Mattern, P.~Schumacher, F.~M. López, M.~C. Raabe, M.~R. Ernst, A.~Aubret, and J.~Triesch, ``{MIMo}: A multi-modal infant model for studying cognitive development,'' 2023.

\bibitem{mujoco}
E.~Todorov, T.~Erez, and Y.~Tassa, ``Mujoco: A physics engine for model-based control,'' in \emph{2012 IEEE/RSJ International Conference on Intelligent Robots and Systems}, 2012, pp. 5026--5033.

\bibitem{schulman2017proximal}
J.~Schulman, F.~Wolski, P.~Dhariwal, A.~Radford, and O.~Klimov, ``Proximal policy optimization algorithms,'' 2017.

\bibitem{mysore2021regularizing}
S.~Mysore, B.~Mabsout, R.~Mancuso, and K.~Saenko, ``Regularizing action policies for smooth control with reinforcement learning,'' in \emph{2021 IEEE International Conference on Robotics and Automation (ICRA)}.\hskip 1em plus 0.5em minus 0.4em\relax IEEE, 2021, pp. 1810--1816.

\end{thebibliography}
\end{document}